\pdfoutput=1

\documentclass[11pt]{article}

\usepackage{EMNLP2023}

\usepackage{times}
\usepackage{latexsym}

\usepackage[T1]{fontenc}

\usepackage[utf8]{inputenc}

\usepackage{microtype}

\usepackage{inconsolata}

\usepackage{adjustbox}
\usepackage{amsmath}
\usepackage{amsfonts}
\usepackage{bbm}
\usepackage{graphicx}
\usepackage{caption}
\usepackage{subcaption}
\usepackage{makecell}
\usepackage{booktabs}
\usepackage{multirow}
\usepackage{xcolor}         

\usepackage{enumitem}
\newcommand{\comment}[1]{}
\usepackage{mathtools}

\newcommand\nnfootnote[1]{%
  \begin{NoHyper}
  \renewcommand\thefootnote{}\footnote{#1}%
  \addtocounter{footnote}{-1}%
  \end{NoHyper}
}

\definecolor{amethyst}{rgb}{0.6, 0.4, 0.8}
\definecolor{applegreen}{rgb}{0.55, 0.71, 0.0}
\definecolor{blue}{rgb}{0.1,0.5,1.0}

\usepackage{hyperref}
\definecolor{Gray}{gray}{0.9}
\definecolor{Red}{rgb}{0.858, 0.08, 0.08}
\definecolor{Green}{rgb}{0.58, 0.958, 0.78}
\hypersetup{
    colorlinks=true,
    citecolor=teal,
    linkcolor=Red,
    urlcolor=Red,
}

\title{Context-dependent Instruction Tuning for Dialogue Response Generation}

\author{Jin Myung Kwak$^1${\color{Red}$^*$}, \: Minseon Kim$^1${\color{Red}$^*$}, \: Sung Ju Hwang$^1$$^,$$^2$ \\  
KAIST$^1$,  DeepAuto$^2$\\
\texttt{\{kwak.jinmyung, minseonkim, sjhwang82\}@kaist.ac.kr}}

\begin{document}
\maketitle
\begin{abstract}
Recent language models have achieved impressive performance in natural language tasks by incorporating instructions with task input during fine-tuning. Since all samples in the same natural language task can be explained with the same task instructions, many instruction datasets only provide a few instructions for the entire task, without considering the input of each example in the task. However, this approach becomes ineffective in complex multi-turn dialogue generation tasks, where the input varies highly with each turn as the dialogue context changes, so that simple task instructions cannot improve the generation performance. To address this limitation, we introduce a context-based instruction fine-tuning framework for each multi-turn dialogue which generates both responses and instructions based on the previous context as input. During the evaluation, the model generates instructions based on the previous context to self-guide the response. The proposed framework produces comparable or even outstanding results compared to the baselines by aligning instructions to the input during fine-tuning with the instructions in quantitative evaluations on dialogue benchmark datasets with reduced computation budget.

\end{abstract}

\nnfootnote{\color{Red}{*} \color{black} Equal contribution; ordering determined by coin toss}

\section{Introduction}
Instruction tuning has significantly enhanced the generalization of pretrained language models across a variety of natural language tasks~\citep{flan21, flan23, selfinstruct, instructgpt4}. Prepending a natural language description of each task to the input during a multi-task learning setting enables the high-performance gain in both seen and unseen tasks. Diverse attempts have been made to design and generate a few instructions for tasks, such as template-based~\citep{flan21}, model-based~\citep{selfinstruct} and human-designed based~\citep{supernaturalinstructions, instructdial}, for instruction tuning. 

In this work, we focus on model-based instruction tuning for a single specific task: multi-turn dialogue response generation. This task requires the generation of diverse responses based on the given input such as dialogue context or additional information (i.e., persona). The existing generic instructions~\citep{instructdial} might be one of the options; however, they easily fail to guide high-quality responses in the multi-turn dialogue as shown in Table~\ref{ablation_table}. This limitation is attributed to the rapid context-switching inherent in dialogue interactions and the intricate conditions of each dialogue.

\begin{figure*}
    \centering
    \includegraphics[width=0.93\textwidth]{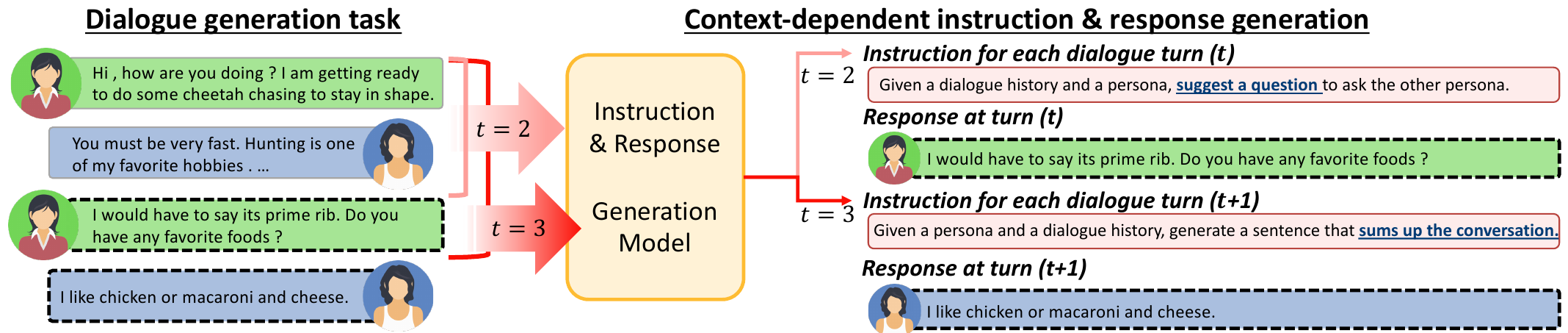}
    \vspace{-0.05in}
    \caption{\textbf{Concept.} Our model generates both instructions and responses at each turn ($t$) based on previous context ($0 \sim t-1$).}
    \vspace{-0.15in}
    \label{fig1:concept}
\end{figure*}
To overcome such a challenge, we propose a novel framework that provides context-dependent instructions for dialogue response generation without any human interference. Since there are no instruction datasets for each multi-turn conversation, we first construct context-based instructions for dialogue generation tasks by training an instruction generation model. To condition the context in generation, given the context-based instructions, our model learns to generate both instructions and responses with the single model using a multi-task learning approach via sentinel tokens. From these multi-task generations, the model is able to generate context-dependent instructions and generate instruction-guided responses afterward, which enhances the performance of dialogue response generation to produce diverse and coherent responses.

For evaluation, we conduct experiments on two benchmark datasets, DailyDialog~\citep{dailydialog} and PersonaChat~\citep{personachat}, for multi-turn dialogue response generation. Our framework achieves outstanding performance in both datasets. Our experimental results emphasize the importance of aligning instructions to the input during instruction tuning for the multi-turn dialogue response generation tasks. Our contributions are as follows:

\vspace{-0.15in}
\begin{itemize}
[itemsep=1.4mm, parsep=0pt, leftmargin=*]
\item We uncover the key insight that utilizing context-dependent instructions significantly improves the performance of multi-turn dialogue generation, compared to relying on a few fixed task instructions.
\item We introduce a novel instruction tuning that generates both instructions and responses as a multi-task generation that is conditioned on the previous dialogue context.
\item We demonstrate comparable or even outstanding results compared to baselines with a lower computation budget in quantitative evaluations on the DailyDialog and PersonaChat datasets.\end{itemize}

\section{Related Work}
\paragraph{Instruction Tuning \& Generation.} Instruction tuning has enabled training powerful and general-purpose language models~\citep{flan21, instructgpt3}. Recent works aim to discover the optimal instructions for each task~\citep{ape,iprompt} or expand the instruction datasets~\citep{selfinstruct}. The former approaches primarily focus on generating a task-representative instruction that could be used in the test time. The latter approach concentrates on generating diverse instructions and examples for various tasks, which are used for training, but not in the test time. Different from the prior works, we generate instructions based on each context of dialogues and use them in generating adequate responses in the conversation. This enables us to generate improved responses which self-guide the generation of responses with the corresponding context-based instructions even during test time. 

\paragraph{Dialogue Generation.} Dialogue generation is a versatile task that can be directly applied to real-world applications to various scenarios, including persona-grounded, knowledge-grounded, goal-oriented, and socially-grounded conversations~\citep{personachat, wow, multiwoz,soda}. However, it is a complex task that requires both natural language understanding and generation at the same time. Many studies explore different approaches such as training objectives~\citep{wu2019guiding, song2020generate}, decoding techniques~\citep{qi2020prophetnet}, architectural improvements~\citep{zheng2020pre}, data augmentation~\citep{rastogi2020towards}, and pretraining or finetuning language models specifically for dialogue tasks~\citep{zhang2019dialogpt}, to enhance the quality of generated responses.

\section{Context-based Instruction and Dialogue Generation}
In this section, we introduce our model that learns to generate both instructions and responses given dialogue context. We first explain the dialogue generation task and its notations. Next, we elaborate on how to generate the instructions for the multi-turn dialogue task. Lastly, we detail our training process of our proposed framework which only uses a single language model. 

\vspace{-0.1in}
\subsection{Dialogue Generation Task}
The goal of dialogue generation models is to generate a response $R$ given a dialogue context $C$. In this paper, we refer to the dialogue context as consisting of several past utterances and additional information such as persona. The dialogue dataset $\mathcal{D}$ consists of $\{C^{n},R^{n}\}^{N}_{n=1}$ pairs, where $N$ is the number of observed examples. The models are optimized to learn conditional likelihood $p(r_t | r_{<t}, C)$, predicting $r_t$, which represents $t$-th token of the response $R$, when $C$ is given.


\vspace{-0.1in}
\subsection{Instruction Data Generation}
Before we delve into explaining our model, it is necessary to outline the preliminary preparation process of constructing instructions for dialogue datasets. Since there are not any instruction datasets for dialogue context that include instructions for each turn within every dialogue context, we train an instruction generation model to gather instructions for our model. To the best of our knowledge, we are the first attempt to form context-based instruction construction for the dialogue generation task.

\vspace{-0.05in}
\paragraph{Dataset.} SELF-INSTRUCT~\citep{selfinstruct}, an automatically generated data from GPT3, which consists of 52k examples, is used for training the instruction generation model which contains $78\%$ non-classification task instructions. Considering the emphasis on dialogue generation, we train the instruction generation model using the dataset primarily composed of diverse non-classification tasks.
\vspace{-0.05in}
\paragraph{Training.} 
We train the instruction generation model using FLAN-T5~\citep{flan21}, which is an instruction-finetuned Transformer based language model, with SELF-INSTRUCT~\citep{selfinstruct}. The training dataset consists of $M$ triplets $\{I^{m},X^{m},Y^{m}\}^{M}_{m=1}$, where $I$ represents the instruction, $X$ is the input context of the task, and $Y$ is the response/answer of the given the input context, respectively. We optimize the instruction generation model to learn conditional probability  $P(i_t|i_{<t},X,Y)$, where $i_t$ represents $t$-th token of the instruction $I$, conditioned on previous instruction tokens $i_{<t}$, input $X$ and output $Y$. 
Inspired by the T5 denoising training objective~\citep{T5}, we utilize the sentient token to represent the instruction and predict the corresponding instruction. This approach transforms the generation process into a fill-in-the-blank problem.
\vspace{-0.05in}
\paragraph{Inference.} 
After training the instruction generation model, we are now capable of producing an instruction set for dialogue datasets. In this paper, we have collected instructions for DailyDialog~\citep{dailydialog} and PersonaChat~\citep{personachat}. The instruction generation model returns the following instructions, given the dialogue context as the input $X$ and the next utterance of the given dialogue context as the output $Y$. The instruction generation process applies not only to specific datasets but also to any dialogue datasets.
\begin{table}[t]\centering
\centering
\begin{adjustbox}{width=\linewidth}
%
\begin{tabular}{l|cccc}
\toprule
&
 \multicolumn{4}{c}{DailyDialog} \\
 Model &
 BLEU-1 &
 BLEU-2 &
 Distinct-1 &
 Distinct-2 \\ 
\midrule[0.5pt]
 \multicolumn{1}{l|}{Original} &
 \multicolumn{1}{c}{0.455} &
 \multicolumn{1}{c}{0.388} &
 \multicolumn{1}{c}{0.039} &
 \multicolumn{1}{c}{0.205}
\\ 
\multicolumn{1}{l|}{InstructDial} &
 \multicolumn{1}{c}{0.457} &
 \multicolumn{1}{c}{0.391} &
 \multicolumn{1}{c}{0.039} &
 \multicolumn{1}{c}{0.206} \\
\multicolumn{1}{l|}{Ours} &
 \multicolumn{1}{c}{\textbf{0.470}} &
 \multicolumn{1}{c}{\textbf{0.400}} &
 \multicolumn{1}{c}{\textbf{0.057}} &
 \multicolumn{1}{c}{\textbf{0.256}} \\
\midrule[0.5pt]

\multicolumn{1}{l|}{Upper Bound} &
 \multicolumn{1}{c}{0.561} &
 \multicolumn{1}{c}{0.489} &
 \multicolumn{1}{c}{0.039} &
 \multicolumn{1}{c}{0.212}\\
\bottomrule
\end{tabular}
\end{adjustbox}
\vspace{-0.05in}
\caption{\textbf{Results on comparing the impact of context-based instruction tuning.} Given the same basis model, FLAN-T5 small,  we compare four different settings: without instructions (Original), with a fixed set of instructions (InstructDial), with generated context-based instructions (Ours), and with ground-truth instruction (Upper Bound).} 
\label{ablation_table}
\vspace{-0.2in}
\end{table}

\begin{table*}[t]\centering
\centering
\begin{adjustbox}{width=0.9\textwidth}
%
\begin{tabular}{l|c|cccc|cccc}
\toprule
 & 
 &
 \multicolumn{4}{c|}{DailyDialog} & 
 \multicolumn{4}{c}{PersonaChat} \\
 Model &
 Parameter &
 BLEU-1 &
 BLEU-2 &
 Distinct-1 &
 Distinct-2&
 BLEU-1 &
 BLEU-2 &
 Distinct-1 &
 Distinct-2 \\ 
\midrule[0.5pt]
Seq2Seq~\cite{seq2seq} &
 - &
 0.336 &
 0.238 &
 0.030 &
 0.128 &
 0.448 &
 0.353 &
 0.004 &
 0.016 \\ 



 
 PLATO~\cite{plato}&
 132M &
 0.397 &
 0.311 &
 \underline{0.054} &
 \textbf{0.291} &
 0.406 &
 0.315 &
 \textbf{0.021} &
 \textbf{0.121} \\
 
ProphetNet~\cite{qi2020prophetnet} &
391M &
 0.443 &
 0.392 &
 0.039 &
 0.211&
 0.466 &
 0.391 &
 0.013 &
 0.075 \\

DialogVED~\cite{dialogved} &
392M &
 \textbf{0.481} &
 \textbf{0.421} &
 0.042 &
 0.232 & 
 \underline{0.482} &
 \textbf{0.399} &
 \underline{0.015} &
 \underline{0.094} \\
\midrule[0.5pt]

Ours &
\textbf{77M} & \underline{0.470}
  & \underline{0.400}
  & \textbf{0.057}
  & \underline{0.256}
  & \textbf{0.496}
  & \textbf{0.399}
  & 0.014
  & 0.090
   \\

\midrule[0.5pt]
Human &
-&
- &
 - &
 0.0377 &
 0.294&
 - &
 - &
 0.070 &
 0.412 \\ 
\bottomrule
\end{tabular}
\end{adjustbox}
\vspace{-0.05in}
\caption{\textbf{Performance of DailyDialog and PersonaChat dataset.} We compare the generation performances, as well as the quality and diversity of baselines, using automatic evaluation metrics. Ours achieves comparable performances compared with the baselines by reducing computational costs significantly (\textbf{Bold}: 1st Rank, \underline{Underline}: 2nd Rank).} 
\vspace{-0.2in}
\label{convai2_table}
\end{table*}

\vspace{-0.1in}
\subsection{Our Framework}
We propose a novel generation framework that generates not only appropriate responses to dialogue contexts, but also context-based instructions for dialogue contexts. To achieve this, we employ multi-task learning to train our generative model, enabling us to learn a generalized representation of the dialogue task where use FLAN-T5~\citep{flan21} as a base model. For the training dataset, we incorporate the generated instructions alongside the original dialogue dataset. Therefore, the number of examples does not increase; it remains the same. 

\vspace{-0.05in}
\paragraph{Training.}
To learn how to generate instructions and responses within our framework, we utilize two sentinel tokens: one for dialogue generation and another for instruction generation. We formulate the inputs and outputs of our framework by considering four cases: (1)~ $p(i_t|i_{<t},R,C)$, instruction prediction when both the dialogue context and response are given, (2)~ $p(r_t|r_{<t},I,C)$ response prediction when both the instruction and dialogue context are given, (3)~ $p(i_t|i_{<t},C)$, instruction prediction when only the dialogue context is given, and (4)~$p(r_t|r_{<t},C)$, response prediction when only the dialogue context is given. We randomly sample input-output pairs and optimize our framework by minimizing the negative log-likelihood losses for each case during training.
\vspace{-0.05in}
\paragraph{Inference.} 
Our framework can generate responses and instructions in two distinct generation methods. The first method, referred to as the naive approach, involves the independent generation of instructions and responses based on the dialogue contexts. The second method employs an iterative process, where one is generated after the other. For the direct generation, we first create instructions using the naive approach. Then, using these initially generated instructions, instruction-guided responses are produced. 

\vspace{-0.05in}
\section{Experiments}
\vspace{-0.05in}
We evaluate our context-based instruction tuning in two benchmark datasets, DailyDialog~\citep{dailydialog}, and PersonaChat~\citep{personachat}. From our observation, we find that context-aligned instruction could much improve the performance of dialogue generation compared to utilizing fixed instruction. To demonstrate the effectiveness of context-based instruction tuning, we also compare our work with dialogue generation baselines, which do not utilize instructions. Our proposed model is fine-tuned using FLAN-T5 small~\citep{flan21}. For further details on training and inference, please refer to Appendix~\ref{appendix:exp_detail}.

\vspace{-0.05in}
\paragraph{Baselines.} Our baselines could be categorized into two types: a model trained with a fixed set of instructions and general dialogue model. \textbf{InstructDial} is a FLAN-T5 model that is trained on a fixed number (3 to 5) of human-labeled instructions~\cite{instructdial}. For the comparison of general dialogue models without instructions, we examine the following approaches: \textbf{Seq2Seq}is a vanilla sequence-to-sequence model fine-tuned from scratch, \textbf{Original} is a FLAN-T5 based model fine-tuned without any modifications, \textbf{PLATO} is a BERT-based model with a latent discrete variable, \textbf{ProphetNet} is a Transformer-based model that predicts future n-grams, \textbf{DialogVED} is a Transformer-based model with a continuous latent variable.

\vspace{-0.05in}
\paragraph{Automatic Evaluation Metrics.} BLEU-1/2 measures the n-gram overlap between the generated responses and the ground-truth responses, while Distinct-1/2 quantifies the diversity of the generated responses. Higher values in both metrics indicate better performance.

\vspace{-0.05in}
\subsection{Experimental results}
\label{result:inst-baseline}
\paragraph{Results on Context-based Instruction Tuning.}The primary motivation for our context-based instruction fine-tuning is the necessity of providing precise guidelines to language models in each turn of conversations. As shown in Table~\ref{ablation_table}, the InstructDial, trained with a fixed generic instruction set, could not much improve the dialogue generation performance compared to the Original, which does not utilize any instructions at all. These results demonstrate the need of effective instructions for dialogue response generation tasks.

Furthermore, assuming that we have access to the future response in the test set, we can generate context-aligned instruction for all test dialogues. When we utilize those instructions, the dialogue generation model could achieve 0.561 performance in the DailyDialog dataset which is the upper bound of instruction tuning. Those instructions are generated by conditioning the ground-truth responses, encapsulating valuable information that strongly guides the model toward generating responses aligned with the ground truth.

However, since the responses are unknown in test time, our model generates context-based instructions depending on the previous conversation. As shown in the results, when the model provided appropriate context-dependent instructions, the model could achieve much-improved performance than fixed instructions.

\vspace{-0.05in}
\paragraph{Comparison on Dialogue Generation Baselines.}
\label{result:dialogue-baseline}
As demonstrated in Table~\ref{convai2_table}, with a straightforward context-based instruction tuning, the model can achieve performance comparable to previous dialogue generation model baselines, but with significantly lower computational costs. While context-dependent instructions enable the model to generate appropriate responses with ease, they may slightly decrease the diversity of responses for each turn. Nevertheless, we maintain that the generation of appropriate responses is the most fundamental aspect that the model should prioritize.
\vspace{-0.05in}
\section{Conclusion}
\vspace{-0.05in}
We introduced a context-based instruction tuning framework for multi-turn dialogue response generation tasks. Utilizing a multi-task generation approach, our approach deploys a single model to generate both instructions and responses via sentinel tokens. It incorporates context information, enabling simultaneous prediction of instructions and future responses. Compared to the traditional fixed set of instruction tuning, our framework allows the model to guide higher-quality responses in conversation more easily by leveraging context-dependent instructions. This introduces a novel perspective for multi-turn dialogue generation tasks, showcasing the potential of context-specific instruction use in dialogue response generation applications.

\section*{Limitations}
The context-dependent instruction tuning for dialogue response certainly facilitates more precise and coherent responses. One could note that it involves a two-step learning process, generating instructions for each context and then mapping these to create responses. However, this aspect should not be perceived as a limitation; rather, it's an integral part of our innovative approach. The requirement to generate instructions for each context might introduce an additional step, but this also plays a significant role in dataset construction, adding value to our method. Hence, while there are additional elements to consider in our approach, they contribute to its uniqueness and efficacy, subtly balancing any perceived disadvantages.





\bibliography{anthology,custom}

\begin{thebibliography}{24}
\expandafter\ifx\csname natexlab\endcsname\relax\def\natexlab#1{#1}\fi

\bibitem[{Bao et~al.(2020)Bao, He, Wang, Wu, and Wang}]{plato}
Siqi Bao, Huang He, Fan Wang, Hua Wu, and Haifeng Wang. 2020.
\newblock \href {https://doi.org/10.18653/v1/2020.acl-main.9} {{PLATO}: Pre-trained dialogue generation model with discrete latent variable}.
\newblock In \emph{Proceedings of the 58th Annual Meeting of the Association for Computational Linguistics}, pages 85--96, Online. Association for Computational Linguistics.

\bibitem[{Budzianowski et~al.(2018)Budzianowski, Wen, Tseng, Casanueva, Ultes, Ramadan, and Ga{\v{s}}i{\'c}}]{multiwoz}
Pawe{\l} Budzianowski, Tsung-Hsien Wen, Bo-Hsiang Tseng, I{\~n}igo Casanueva, Stefan Ultes, Osman Ramadan, and Milica Ga{\v{s}}i{\'c}. 2018.
\newblock \href {https://doi.org/10.18653/v1/D18-1547} {{M}ulti{WOZ} - a large-scale multi-domain {W}izard-of-{O}z dataset for task-oriented dialogue modelling}.
\newblock In \emph{Proceedings of the 2018 Conference on Empirical Methods in Natural Language Processing}, pages 5016--5026, Brussels, Belgium. Association for Computational Linguistics.

\bibitem[{Chen et~al.(2022)Chen, Gong, Wang, Yao, Qi, Wei, Hu, Zhou, Mao, Chen, Cheng, and Duan}]{dialogved}
Wei Chen, Yeyun Gong, Song Wang, Bolun Yao, Weizhen Qi, Zhongyu Wei, Xiaowu Hu, Bartuer Zhou, Yi~Mao, Weizhu Chen, Biao Cheng, and Nan Duan. 2022.
\newblock \href {https://doi.org/10.18653/v1/2022.acl-long.333} {{D}ialog{VED}: A pre-trained latent variable encoder-decoder model for dialog response generation}.
\newblock In \emph{Proceedings of the 60th Annual Meeting of the Association for Computational Linguistics (Volume 1: Long Papers)}, pages 4852--4864, Dublin, Ireland. Association for Computational Linguistics.

\bibitem[{Dinan et~al.(2019)Dinan, Roller, Shuster, Fan, Auli, and Weston}]{wow}
Emily Dinan, Stephen Roller, Kurt Shuster, Angela Fan, Michael Auli, and Jason Weston. 2019.
\newblock {W}izard of {W}ikipedia: Knowledge-powered conversational agents.
\newblock In \emph{Proceedings of the International Conference on Learning Representations (ICLR)}.

\bibitem[{Gupta et~al.(2022)Gupta, Jiao, Yeh, Mehri, Eskenazi, and Bigham}]{instructdial}
Prakhar Gupta, Cathy Jiao, Yi-Ting Yeh, Shikib Mehri, Maxine Eskenazi, and Jeffrey Bigham. 2022.
\newblock \href {https://aclanthology.org/2022.emnlp-main.33} {{I}nstruct{D}ial: Improving zero and few-shot generalization in dialogue through instruction tuning}.
\newblock In \emph{Proceedings of the 2022 Conference on Empirical Methods in Natural Language Processing}, pages 505--525, Abu Dhabi, United Arab Emirates. Association for Computational Linguistics.

\bibitem[{Kim et~al.(2022)Kim, Hessel, Jiang, West, Lu, Yu, Zhou, Bras, Alikhani, Kim, Sap, and Choi}]{soda}
Hyunwoo Kim, Jack Hessel, Liwei Jiang, Peter West, Ximing Lu, Youngjae Yu, Pei Zhou, Ronan~Le Bras, Malihe Alikhani, Gunhee Kim, Maarten Sap, and Yejin Choi. 2022.
\newblock Soda: Million-scale dialogue distillation with social commonsense contextualization.
\newblock \emph{ArXiv}, abs/2212.10465.

\bibitem[{Li et~al.(2017)Li, Su, Shen, Li, Cao, and Niu}]{dailydialog}
Yanran Li, Hui Su, Xiaoyu Shen, Wenjie Li, Ziqiang Cao, and Shuzi Niu. 2017.
\newblock \href {https://aclanthology.org/I17-1099} {{D}aily{D}ialog: A manually labelled multi-turn dialogue dataset}.
\newblock In \emph{Proceedings of the Eighth International Joint Conference on Natural Language Processing (Volume 1: Long Papers)}, pages 986--995, Taipei, Taiwan. Asian Federation of Natural Language Processing.

\bibitem[{Longpre et~al.(2023)Longpre, Hou, Vu, Webson, Chung, Tay, Zhou, Le, Zoph, Wei et~al.}]{flan23}
Shayne Longpre, Le~Hou, Tu~Vu, Albert Webson, Hyung~Won Chung, Yi~Tay, Denny Zhou, Quoc~V Le, Barret Zoph, Jason Wei, et~al. 2023.
\newblock The flan collection: Designing data and methods for effective instruction tuning.
\newblock \emph{arXiv preprint arXiv:2301.13688}.

\bibitem[{Ouyang et~al.(2022)Ouyang, Wu, Jiang, Almeida, Wainwright, Mishkin, Zhang, Agarwal, Slama, Ray et~al.}]{instructgpt3}
Long Ouyang, Jeffrey Wu, Xu~Jiang, Diogo Almeida, Carroll Wainwright, Pamela Mishkin, Chong Zhang, Sandhini Agarwal, Katarina Slama, Alex Ray, et~al. 2022.
\newblock Training language models to follow instructions with human feedback.
\newblock \emph{Advances in Neural Information Processing Systems}, 35:27730--27744.

\bibitem[{Peng et~al.(2023)Peng, Li, He, Galley, and Gao}]{instructgpt4}
Baolin Peng, Chunyuan Li, Pengcheng He, Michel Galley, and Jianfeng Gao. 2023.
\newblock Instruction tuning with gpt-4.
\newblock \emph{arXiv preprint arXiv:2304.03277}.

\bibitem[{Qi et~al.(2020)Qi, Yan, Gong, Liu, Duan, Chen, Zhang, and Zhou}]{qi2020prophetnet}
Weizhen Qi, Yu~Yan, Yeyun Gong, Dayiheng Liu, Nan Duan, Jiusheng Chen, Ruofei Zhang, and Ming Zhou. 2020.
\newblock Prophetnet: Predicting future n-gram for sequence-to-sequence pre-training.
\newblock \emph{arXiv preprint arXiv:2001.04063}.

\bibitem[{Raffel et~al.(2020)Raffel, Shazeer, Roberts, Lee, Narang, Matena, Zhou, Li, and Liu}]{T5}
Colin Raffel, Noam Shazeer, Adam Roberts, Katherine Lee, Sharan Narang, Michael Matena, Yanqi Zhou, Wei Li, and Peter~J. Liu. 2020.
\newblock \href {http://jmlr.org/papers/v21/20-074.html} {Exploring the limits of transfer learning with a unified text-to-text transformer}.
\newblock \emph{Journal of Machine Learning Research}, 21(140):1--67.

\bibitem[{Rastogi et~al.(2020)Rastogi, Zang, Sunkara, Gupta, and Khaitan}]{rastogi2020towards}
Abhinav Rastogi, Xiaoxue Zang, Srinivas Sunkara, Raghav Gupta, and Pranav Khaitan. 2020.
\newblock Towards scalable multi-domain conversational agents: The schema-guided dialogue dataset.
\newblock In \emph{Proceedings of the AAAI Conference on Artificial Intelligence}, pages 8689--8696.

\bibitem[{Singh et~al.(2023)Singh, Morris, Aneja, Rush, and Gao}]{iprompt}
Chandan Singh, John~Xavier Morris, Jyoti Aneja, Alexander~M Rush, and Jianfeng Gao. 2023.
\newblock \href {https://openreview.net/forum?id=GvMuB-YsiK6} {Explaining patterns in data with language models via interpretable autoprompting}.

\bibitem[{Song et~al.(2020)Song, Wang, Zhang, Liu, and Liu}]{song2020generate}
Haoyu Song, Yan Wang, Wei-Nan Zhang, Xiaojiang Liu, and Ting Liu. 2020.
\newblock Generate, delete and rewrite: A three-stage framework for improving persona consistency of dialogue generation.
\newblock \emph{ACL}.

\bibitem[{Vinyals and Le(2015)}]{seq2seq}
Oriol Vinyals and Quoc Le. 2015.
\newblock A neural conversational model.
\newblock \emph{arXiv preprint arXiv:1506.05869}.

\bibitem[{Wang et~al.(2022{\natexlab{a}})Wang, Kordi, Mishra, Liu, Smith, Khashabi, and Hajishirzi}]{selfinstruct}
Yizhong Wang, Yeganeh Kordi, Swaroop Mishra, Alisa Liu, Noah~A. Smith, Daniel Khashabi, and Hannaneh Hajishirzi. 2022{\natexlab{a}}.
\newblock Self-instruct: Aligning language model with self generated instructions.

\bibitem[{Wang et~al.(2022{\natexlab{b}})Wang, Mishra, Alipoormolabashi, Kordi, Mirzaei, Arunkumar, Ashok, Dhanasekaran, Naik, Stap et~al.}]{supernaturalinstructions}
Yizhong Wang, Swaroop Mishra, Pegah Alipoormolabashi, Yeganeh Kordi, Amirreza Mirzaei, Anjana Arunkumar, Arjun Ashok, Arut~Selvan Dhanasekaran, Atharva Naik, David Stap, et~al. 2022{\natexlab{b}}.
\newblock Super-naturalinstructions:generalization via declarative instructions on 1600+ tasks.
\newblock In \emph{EMNLP}.

\bibitem[{Wei et~al.(2022)Wei, Bosma, Zhao, Guu, Yu, Lester, Du, Dai, and Le}]{flan21}
Jason Wei, Maarten Bosma, Vincent Zhao, Kelvin Guu, Adams~Wei Yu, Brian Lester, Nan Du, Andrew~M. Dai, and Quoc~V Le. 2022.
\newblock \href {https://openreview.net/forum?id=gEZrGCozdqR} {Finetuned language models are zero-shot learners}.
\newblock In \emph{International Conference on Learning Representations}.

\bibitem[{Wu et~al.(2020)Wu, Li, Wang, Chen, Wong, Feng, Huang, and Wang}]{wu2019guiding}
Bowen Wu, MengYuan Li, Zongsheng Wang, Yifu Chen, Derek Wong, Qihang Feng, Junhong Huang, and Baoxun Wang. 2020.
\newblock Guiding variational response generator to exploit persona.
\newblock \emph{ACL}.

\bibitem[{Zhang et~al.(2018)Zhang, Dinan, Urbanek, Szlam, Kiela, and Weston}]{personachat}
Saizheng Zhang, Emily Dinan, Jack Urbanek, Arthur Szlam, Douwe Kiela, and Jason Weston. 2018.
\newblock \href {https://doi.org/10.18653/v1/P18-1205} {Personalizing dialogue agents: {I} have a dog, do you have pets too?}
\newblock In \emph{Proceedings of the 56th Annual Meeting of the Association for Computational Linguistics (Volume 1: Long Papers)}, pages 2204--2213, Melbourne, Australia. Association for Computational Linguistics.

\bibitem[{Zhang et~al.(2020)Zhang, Sun, Galley, Chen, Brockett, Gao, Gao, Liu, and Dolan}]{zhang2019dialogpt}
Yizhe Zhang, Siqi Sun, Michel Galley, Yen-Chun Chen, Chris Brockett, Xiang Gao, Jianfeng Gao, Jingjing Liu, and Bill Dolan. 2020.
\newblock Dialogpt: Large-scale generative pre-training for conversational response generation.
\newblock \emph{ACL}.

\bibitem[{Zheng et~al.(2020)Zheng, Zhang, Huang, and Mao}]{zheng2020pre}
Yinhe Zheng, Rongsheng Zhang, Minlie Huang, and Xiaoxi Mao. 2020.
\newblock A pre-training based personalized dialogue generation model with persona-sparse data.
\newblock In \emph{Proceedings of the AAAI Conference on Artificial Intelligence}, pages 9693--9700.

\bibitem[{Zhou et~al.(2023)Zhou, Muresanu, Han, Paster, Pitis, Chan, and Ba}]{ape}
Yongchao Zhou, Andrei~Ioan Muresanu, Ziwen Han, Keiran Paster, Silviu Pitis, Harris Chan, and Jimmy Ba. 2023.
\newblock \href {https://openreview.net/forum?id=92gvk82DE-} {Large language models are human-level prompt engineers}.
\newblock In \emph{The Eleventh International Conference on Learning Representations}.

\end{thebibliography}
\bibliographystyle{acl_natbib}
\appendix
\section{Appendix}
\label{sec:appendix}

\subsection{Datasets}
\paragraph{SELF INSTRUCT~\citep{selfinstruct}.}
The SELF INSTRUCT dataset, generated by a GPT3 language model, comprises 52k instructions matched with 82k input-output instances. There are 11,584 instructions for classification tasks, and 40,861 instructions for non-classification tasks. Average length of the instruction is 15.9 words.

\paragraph{DailyDialog~\citep{dailydialog}.}
DailyDialog is a dialogue dataset consisting of conversations between two individuals. It includes 13,118 multi-turn dialogues, with an average of 7.9 speaker turns per dialogue and 14.6 tokens per utterance. The dataset is partitioned into an 11,118-dialogue training set, and validation and test sets, each containing 1,000 dialogues.

\paragraph{PersonaChat~\citep{personachat}.}
PersonaChat, sourced from Amazon Mechanical Turk, includes 955 training personas, 100 validation personas, and 100 test personas. Utilizing these personas, 10,907 dialogues were generated, constituting a total of 162,064 conversation samples. The primary task objective is to generate appropriate responses, considering both the conversational context and the given persona.

\subsection{Experimental Details}
\paragraph{Modeling Details}
\label{appendix:exp_detail}
We use FLAN-T5 as a base model. Model is trained with 512 max length, 32 batch sizes per GPU, $5e^{-4}$ learning rate, and 40 epochs. The hyperparameters are empirically determined. For all datasets, we set AdamW optimizer with $1e^{-8}$ epsilon and a linear scheduler with warmup ratio 0.03 and weight decay $1e^{-6}$. Trainings are performed on a 1 NVIDIA RTX 3090 for approx. 6 hours on DailyDialog and 0 hours on PersonaChat.

For response generation, the max length of a generation is set to 128. All the generations use beam search with a beam size of 3 and no repeat n-gram size of 3 is applied.

\subsection{Generation Examples}
The Table~\ref{table:appendix_daily} and Table~\ref{table:appendix_persona} are the examples of generated instructions and response from our model for DailyDialog and PersonaChat.

\begin{table*}[t]\centering

\caption{\textbf{Example responses and instructions that are generated from our model in DailyDialog.} Context contains previous context or previous dialogue. Reply is provided previous response to the context. Generations are responses to the reply. instructions are the context-dependent instructions to generate more coherent responses.}

\begin{adjustbox}{width=\textwidth}
\begin{tabular}{l|l}
\toprule
  
  Context&  "We've managed to reduce our energy consumption in our factory by about 15 per cent \\&\quad in the last two years."\\
                   & "That's excellent. How have you managed that?"\\
  Instruction & Given a context, generate a dialogue between two people.
\\
  Response & We haven't had a chance to reduce our energy consumption in our factory, but we've \\&\quad managed to reduce it by about 15 per cent.\\
  \midrule
  Context& "What a nice day!"\\
            &"yes. How about going out and enjoying the sunshine on the grass?"\\
            &"great, let's go!"\\
            &"hey, darling, I think I might have a little heatstroke from being in the sun all day. I was \\&\quad so relaxed. It felt as if I were in another world."\\
            &"exactly. You know, the sunshine and wind remind me of our honeymoon. You remember? \\&\quad The island, the sound of the waves, the salty sea air and the sunshine..."\\
            &"yes, it was wonderful but it's already been a year. How time flies!"\\
            &"why not go again to celebrate out one-year anniversary? We can go to the same beach, \\&\quad stay in the same hotel and enjoy a dinner in the same restaurant."\\
            &"are you kidding? Can you afford it? Do you think we can get a room on such short notice?"\\
            &"never mind that, I'll take care of it. Are you available next week?"\\
            &"yeah, I think so."\\
  Instruction & Generate a final sentence for the following dialogue.
 \\
  Response & I'll give you a call next week, so I can put you on a waiting list.
 \\
  \midrule
  Context&                     "Frank's getting married, do you believe this?"\\
                    &"Is he really?"\\
                    &"Yes, he is. He loves the girl very much."\\
                    &"Who is he marring?"\\
                    &"A girl he met on holiday in Spain, I think."\\
                    &"Have they set a date for the wedding?"\\
  Instruction & Given a context, generate a response from person 1 to person 2.
\\
  Response & Is he really?\\
  \midrule
  Context& "Do you have any hobbies?" \\
        &"Yes, I like collecting things." \\
  Instruction & Given a context, generate a question that can be asked.

\\
  Response & What kind of things do you collect?
 \\
  \midrule
  Context&  "OK. Come back into the classroom, class."\\
                    &"Does the class start again, Mam?" \\
                    &"Yes, but to the heath, this break has long enough." \\
                    &"Oh, breaktime flies."\\
                    &"Come on, baby. It's time now."\\
                    &"Mam, another minute, could I?" \\
  Instruction & Given a context, generate a funny line.

 \\
  Response & Come on, baby. It's time to breathe.

\\
  \bottomrule
\end{tabular}
\label{table:appendix_daily}
\end{adjustbox}

\end{table*}
\begin{table*}[t]\centering

\caption{\textbf{Example responses and instructions that are generated from our model in PersonaChat.} Context contains previous context or previous dialogue. Reply is provided previous response to the context. Generations are responses to the reply. instructions are the context-dependent instructions to generate more coherent responses.}

\begin{adjustbox}{width=\textwidth}
\begin{tabular}{l|l}
\toprule
  Context& "Do you live in a house or apartment ?"\\
            &"We watched a show about animals like him"\\
            &"I love those shows . I am really craving cake ."\\
            &"Why does that matter any ? I went outdoors to play frisbee"\\
            &"It matters because i have a sweet tooth ."\\
  Persona & 
                    "I love to meet new people .",
                    "I have a turtle named timothy ."\\
  Instruction & Generate a funny joke in the given context.\\
  Response & Lol What is your favorite sweet tooth?\\
  \midrule
  Context& "Hey there . how are you ?"\\
        &"Good , do you like animals ?"\\
         &"Yeah , I like cats . I have one ."\\
  Persona & "I am afraid of heights .",
                    "I love animals and have two dogs ."\\
  Instruction & Given a persona and a context, generate a response that is appropriate for the persona.\\
  Response & Me, too. I have a dog. \\
  \midrule
  Context& "I am a chef and I can cook good but i am not happy about my weight ."\\
            &"Oh a chef can you make me a cheese pizza ? I am starving !"\\
            &"Short hair is ok for guys and dogs but my husband hates it on women ."\\
            &"I have short hair but i may grow it out for my fave holiday christmas ."\\
Persona & "I have three dogs .",
                    "My favorite holiday is christmas .",\\
  Instruction & Given a persona and a context, generate a dialogue that is appropriate for the persona.\\
  Response & Oh i see. Do you have any holidays coming up? I love christmas.\\
  \midrule
  Context& "I like to travel as well . i am going to budapest on an acting assignment ."\\
                    &"How fun ! what a great place to take pictures !"\\
                    &"I eat sardines every morning to keep my weight down , no sweets ."\\
                    &"Woah ! salty ! i bake every weekend , so luckily i work out regularly !"\\
                    &"I am studying for my ged so i bring my books on the trade mill ."\\
Persona & "I just had surgery .",
                    "I never finished high school ."\\
  Instruction & Add a new sentence to the following conversation.\\
  Response & I have never finished high school, so I do not have much time to bake.\\
  \midrule
  Context& "Nice ! I just finished feeding my fish . . . I collect exotic fish btw"\\
                    &"Very cool ! My daughter wants a fish tank !"\\
                    &"Are you a teacher ? I go to school myself . . . for hairdressing"\\
                    &"I am a literature teacher ! Nothing like a good book ! nice ! !"\\
Persona & "I have two sons .",
                    "I eat granola for breakfast every day .",\\
  Instruction & Given a persona and a context, generate a response to the following conversation.\\
  Response & No, I am a stay at home mom.\\
  \bottomrule
\end{tabular}
\label{table:appendix_persona}
\end{adjustbox}

\end{table*}

\end{document}